%% file: main.tex
\documentclass[10pt,twocolumn,letterpaper]{article}

\usepackage{cvpr}              % To produce the CAMERA-READY version
\usepackage[table]{xcolor}

\definecolor{cvprblue}{rgb}{0.21,0.49,0.74}
\usepackage[pagebackref,breaklinks,colorlinks,allcolors=cvprblue]{hyperref}
\def\paperID{38819} % *** Enter the Paper ID here
\def\confName{CVPR}
\def\confYear{2026}

\title{GenTract: Generative Global Tractography}

\author{Alec Sargood$^{1}$ \quad Lemuel Puglisi$^{2}$ \quad Elinor Thompson$^{1}$ \quad Mirco Musolesi$^{3}$ \quad Daniel C. Alexander$^{1}$\\
$^{1}$Hawkes Institute and Department of Computer Science, University College London, UK\\
$^{2}$Department of Maths and Computer Science, University of Catania, Italy\\
$^{3}$AI Centre and Department of Computer Science, University College London, UK\\
{\tt\small alec.sargood.23@ucl.ac.uk}
}

\begin{document}
\maketitle
\input{sec/0_abstract}

\begin{figure}[t]
  \centering
  \includegraphics[width=0.8\linewidth]{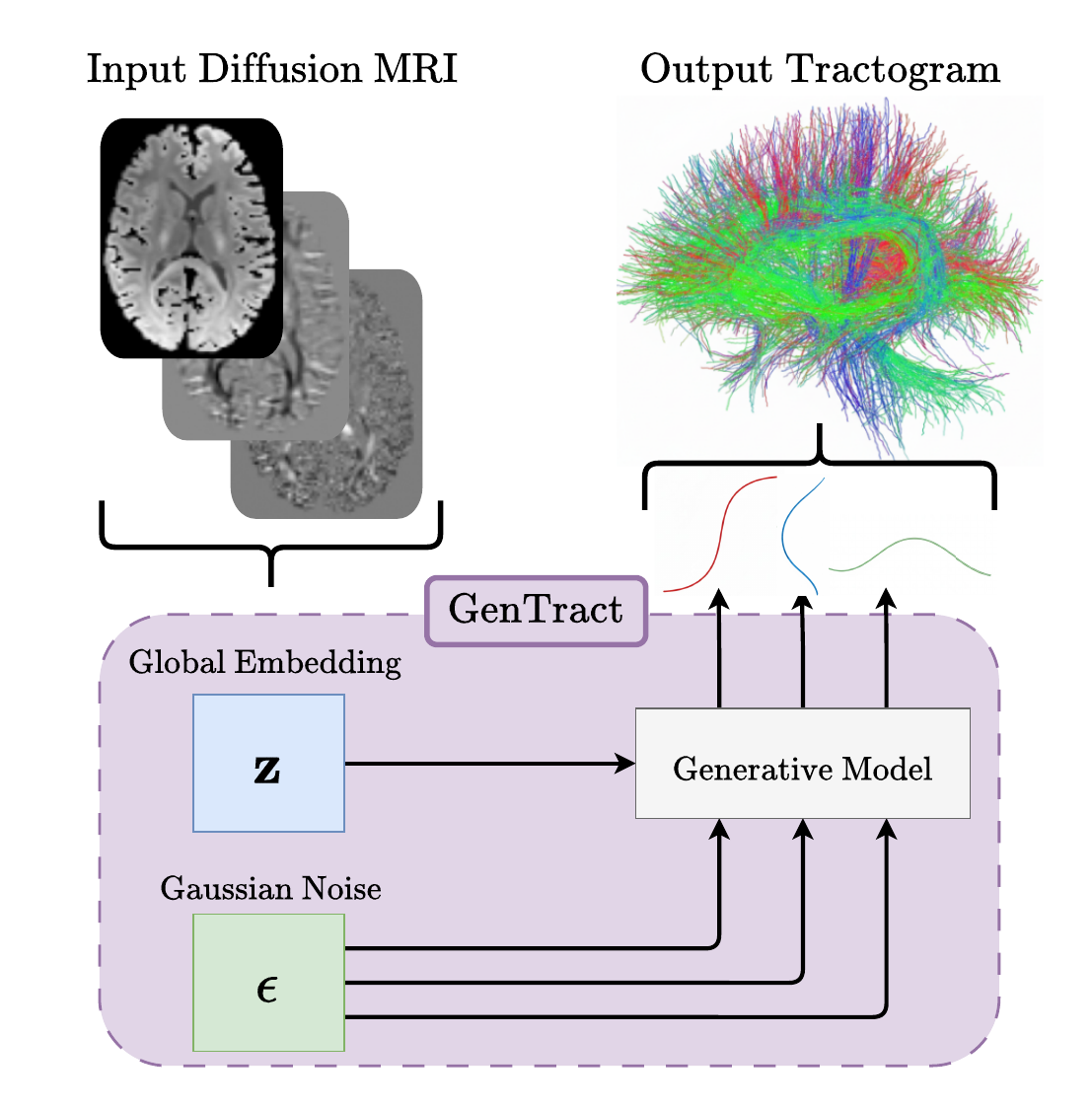}
  \caption{Overview of our proposed GenTract methodology. We provide a learned embedding of, $\mathbf{z}$, representing whole dMRI information. Starting from Gaussian noise ($\epsilon$), all coordinates of individual streamlines are generated in parallel. Generated streamlines are collated to form the output tractogram.}
  \label{fig:intro_figure}
\end{figure}

\input{sec/1_intro}
\input{sec/2_prelims}
\input{sec/3_methods}
\input{sec/4_experiments}
\input{sec/5_discussion}
{
    \small
    \bibliographystyle{ieeenat_fullname}
    \bibliography{main}
}

% WARNING: do not forget to delete the supplementary pages from your submission
\appendix
\input{sec/X_suppl}

\end{document}

%% file: sec/0_abstract.tex
\begin{abstract}
Tractography is the process of inferring the trajectories of white-matter pathways in the brain from diffusion magnetic resonance imaging (dMRI). Local tractography methods, which construct streamlines by following local fiber orientation estimates stepwise through an image, are prone to error accumulation and high false positive rates, particularly on noisy or low-resolution data. In contrast, global methods, which attempt to optimize a collection of streamlines to maximize compatibility with underlying fiber orientation estimates, are computationally expensive. To address these challenges, we introduce GenTract, the first generative model for global tractography. We frame tractography as a generative task, learning a direct mapping from dMRI to complete, anatomically plausible streamlines. We compare both diffusion-based and flow matching paradigms and evaluate GenTract’s performance against state-of-the-art baselines. Notably, GenTract achieves precision 1.8$\times$ and 2.1$\times$ higher than the next-best methods, DDTracking and TractOracle, respectively. This advantage becomes even more pronounced in challenging low-resolution and noisy settings, where it outperforms the closest competitor by a factor of 3.5. By producing tractograms with high precision on research-grade data while also maintaining reliability on imperfect, lower-resolution data, GenTract represents a promising solution for global tractography.
\end{abstract}

%% file: sec/1_intro.tex
\section{Introduction}
\label{sec:intro}

The intricate network of white matter pathways forming the human connectome provides the structural architecture for all neural communication~\citep{filley2016white}. The ability to map these connections in vivo with diffusion-weighted Magnetic Resonance Imaging (dMRI) has offered insights into brain development~\citep{dubois2014early}, aging~\citep{liu2017aging}, and the pathophysiology of neurological disorders~\citep{fields2008white}, while also serving as a tool for neurosurgical planning~\citep{kamagata2024advancements}. However, current methods to reconstruct these pathways, known as tractography, face fundamental challenges that have limited their accuracy and reliability~\citep{mangin2013toward,maier2017challenge}.

Tractography algorithms use information from a dMRI scan, which comprises a set of Diffusion Weighted Images (DWIs), each sensitive to water diffusion in a distinct direction. Because water diffuses more readily along brain fibers than across them, the full set of DWIs can be used to estimate the distribution of fiber orientations within each voxel. The dominant tractography paradigm, local tracking~\cite{descoteaux2008deterministic,tournier2010improved,li2025ddtracking,theberge2024tractoracle,neher2017fiber}, uses these voxel-wise orientations to trace the trajectory of the white matter fibers step-by-step in the form of `streamlines': lines in 3D space that represent the best estimates of the underlying fiber trajectories. However, global connectivity mapping using the local search strategy makes these methods prone to error accumulation, resulting in high false-positive connections~\citep{maier2017challenge,schilling2022prevalence}. This problem is especially pronounced in clinical dMRI scans, which are often limited to acquisition times of a few minutes leading to lower spatial resolution and signal-to-noise ratio compared to the longer, high-resolution protocols used in research. In addition, these methods depend on heterogeneous preprocessing pipelines, especially for constructing seeding masks that define streamline initiation points. This process, which may involve anything from manual segmentation to thresholding of tissue masks, introduces substantial operator- and method-dependent variability, undermining reproducibility~\citep{kruper2021evaluating,theberge2024tractoracle}.

Global tractography~\cite{daducci2025global} seeks to address these limitations by inferring the most probable streamline configurations across the entire brain simultaneously through a single optimization problem. Compared with local methods, global approaches reduce dependence on seeding masks and leverage broader anatomical context. However, existing global approaches depend on traditional optimization schemes that are computationally inefficient, susceptible to suboptimal solutions, and often fail to generate complete tractograms, thereby limiting their practical utility~\citep{maier2017challenge,jeurissen2019diffusion,rheault2025current}.

\paragraph{Contributions.} In this paper, we introduce \textbf{GenTract} (\textbf{Gen}erative \textbf{Tract}ography), reframing tractography as a generative task, conditioned on global information, as illustrated in Figure~\ref{fig:intro_figure}. GenTract is designed to resolve the limitations of both paradigms: it tackles the reliance of local tracking on step-wise information, while solving the computational inefficiency and optimization failures of existing global methods. Our primary contributions are:

\begin{enumerate}
    \item We are the first to formulate tractography as a global, conditional generative task conditioned on the whole set of fiber orientation distribution functions (fODFs) for a given individual. GenTract treats tractography as a sampling process, where the model draws streamlines from a learned conditional distribution, producing all streamline coordinates simultaneously. This overcomes the sequential error propagation of step-wise methods, and also eliminates the need for a seeding mask, reducing method-dependent variability.
    
    \item We propose a modular generative architecture for global tractography compatible with multiple generative paradigms. We provide the first implementation and analysis of both Diffusion and Flow Matching for the task of global tractography, establishing the effectiveness of both schemes.

    \item We conduct a comparative analysis of GenTract against traditional and state-of-the-art (SOTA) tractography paradigms. This analysis demonstrates that GenTract achieves state-of-the-art level of precision on high-fidelity data while exhibiting greater robustness in noisy, low-resolution settings.
    
\end{enumerate}

\section{Related Works}

\paragraph{Local Methods.}\label{sec:localmethods} 
Classical local tractography methods, such as SD Stream~\cite{descoteaux2008deterministic} and iFOD2~\cite{tournier2010improved}, track streamlines by using voxel-wise diffusion information to take the next tracking step. However, they break down in complex fiber configurations like crossing or kissing `bundles'~\cite{rheault2025current}: collections of fibers that form neuroanatomical pathways, connecting two regions of the brain. Machine learning (ML) based techniques have increasingly advanced the field of tractography, improving accuracy over classical methods~\cite{zhang2025think}. For example, the authors in~\cite{neher2017fiber} propose the first ML-based approach, employing random forest classifiers to model the relationship between local image data and sequential tracking steps. More recently, TrackToLearn~\cite{theberge2021track} reframes tractography as a reinforcement learning problem, where agents start from seed points and are guided by rewards that encourage geometrically plausible trajectories and alignment with the diffusion signal. Despite being able to produce more complete connectivity mappings, these methods still produce many false positives, especially in low-resolution data where blurred fiber orientations cause tracking drift. To address this issue, TractOracle~\cite{theberge2024tractoracle} builds on TrackToLearn by incorporating anatomical priors into the reward function, providing agents with additional contextual information, aiding tracking and lowering false positives. Similarly, DDTracking~\cite{li2025ddtracking}, a supervised diffusion-based tracking algorithm, uses neighboring voxel information to generate more reliable next-step predictions. However, these methods still propagate streamlines in an autoregressive manner and are thus limited in their robustness to underlying data perturbations.

\paragraph{Global Methods.}\label{sec:globalmethods} 
Existing global tractography methods~\cite{daducci2025global,jbabdi2007bayesian} address the error accumulation of local methods by treating fiber reconstruction as a single inverse problem, optimizing the entire whole-brain fiber configuration simultaneously to find the solution that best explains the underlying diffusion data. This concept is introduced in \cite{mangin2002framework}, which models short fiber segments as `spins' and finds the lowest-energy state for the system. This is improved by \cite{fillard2009novel}, where these segments can move and rotate to better fit complex crossings. A major limitation is computation time, which is addressed by the authors in \cite{reisert2011global}, who introduce an efficient algorithm that makes global reconstruction more practical by optimizing a global energy function using a faster particle-based approach. The current standard, tckglobal \cite{christiaens2015global}, further refines this by integrating a multi-tissue model. However, these classical optimization approaches still remain a significant computational bottleneck, limiting their application for generating dense whole-brain tractograms. Furthermore, they can exhibit optimization failures, and get stuck in sub-optimal solutions, causing segments to incorrectly connect or terminate prematurely~\cite{christiaens2015global,rheault2025current}. As such, although more robust to local perturbations, global methods have not been adopted as mainstream approaches for tractography. GenTract leverages deep generative modeling to learn a conditional distribution over streamlines based on the entire diffusion signal. This approach maintains the robustness of classical global methods while replacing their costly optimization with a generative process that produces all streamline coordinates in parallel. This process is guided by context from the entire dMRI volume, leading to improvements in both precision and efficiency.

%% file: sec/2_prelims.tex
\section{Preliminaries}
\label{sec:prelims}

\textbf{dMRI Representations.}\label{sec:fodf} Most modern tractography algorithms rely on per-voxel estimates of the fODF for tracking. The fODF, $f(\theta,\phi)$, is a continuous function on the sphere, defined by polar angle $\theta$ and azimuthal angle $\phi$, that represents the distribution of fiber orientations within the voxel. It is designed to capture complex fiber configurations, and is typically estimated from the raw DWIs. For computational tractability, the continuous fODF in each voxel is projected onto an orthogonal Spherical Harmonic (SH) basis, $Y_l^k(\theta,\phi)$:
\begin{equation}
f(\theta,\phi) \approx \sum_{l=0}^{L_{\text{max}}} \sum_{k=-l}^{l} \alpha_{lk} Y_l^k(\theta,\phi).
\end{equation}
Here, $l$ and $k$ are the SH degree and order, and $L_{\text{max}}$ is the maximum harmonic order (e.g., a common choice of $L_{\text{max}}=6$ yields 28 coefficients). This process yields a set of SH coefficients, $\alpha_{lk}$, which provide a compact representation of the fODF for that voxel. The input to tractography is a 4D tensor of size $H \times W \times D \times m$, where each spatial location $(H, W, D)$ contains a vector of $m$ SH coefficients (e.g., $m = 28$ for $L_{\text{max}} = 6$).

\paragraph{Diffusion and Flow Matching Models.} Modern generative models learn to reverse a continuous-time process mapping between data and noise. Diffusion Models (DMs)~\cite{ho2020denoising} learn to reverse a forward noising process (often an SDE) that transforms data $x_0$ into noise $\epsilon \sim \mathcal{N}(0, I)$. Generation involves solving the learned reverse-time SDE. Training is done by learning to predict the added noise $\epsilon$ from a corrupted sample $x_t = \alpha_t x_0 + \sigma_t \epsilon$ using a network $\epsilon_\theta(x_t, t)$:
\begin{equation}\label{eq:dmeqn}
\mathcal{L}_{\text{D}}(\theta) = \mathbb{E}_{t, x_0, \epsilon} \left[ \| \epsilon_\theta(x_t, t) - \epsilon \|^2 \right].
\end{equation}
In contrast, Flow Matching (FM) models~\cite{lipman2022flow} directly learn the deterministic vector field $\mathbf{v}_\theta$ of an ordinary differential equation (ODE), $\frac{dx_t}{dt} = \mathbf{v}_\theta(x_t, t)$, which transports mass from a noise distribution $x_0$ to the data distribution $x_1$. For a simple linear interpolant $x_t = (1-t)x_0 + t x_1$, the target field is $\mathbf{v}=x_1-x_0$. The model is trained via a direct regression loss:
\begin{equation}\label{eq:fm_loss}
\mathcal{L}_{\text{FM}}(\theta) = \mathbb{E}_{t, x_0, x_1} \left[ \| \mathbf{v}_\theta(x_t, t) - \mathbf{v} \|^2 \right],
\end{equation}
and inference is performed by solving the learned ODE, starting from $x_0$.

%% file: sec/3_methods.tex
\section{Methodology: GenTract}
\label{sec:methods}

\begin{figure*}[t]
\centering
\includegraphics[width=0.95\linewidth]{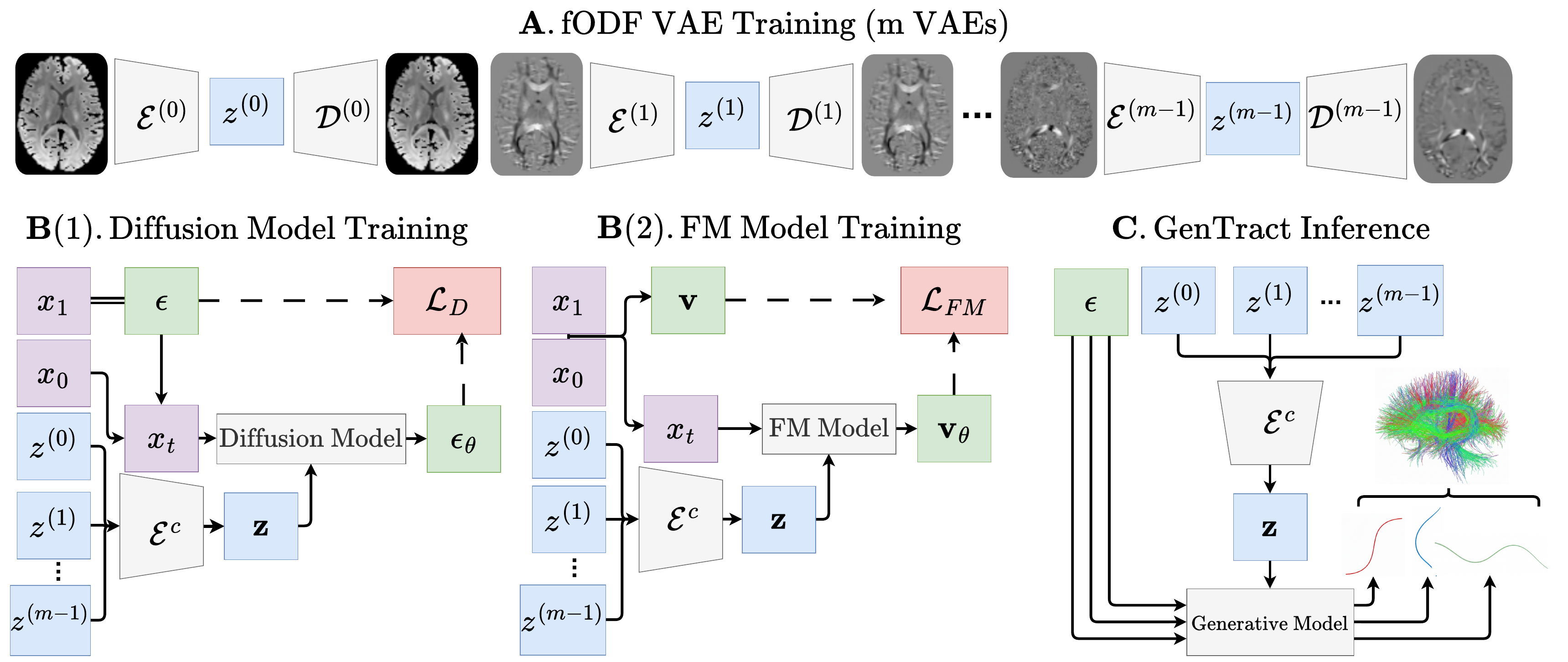}
\caption{Overview of the GenTract framework. \textbf{A}: VAEs encode fODF coefficients into latent representations. \textbf{B(1)} and \textbf{B(2)}: The training protocol for Diffusion and FM models respectively, using the learned $z^{(i)}$ as input. Losses are back-propagated through both the generative model and the class-conditioned encoder $\mathcal{E}^c$. \textbf{C}: The inference process, where streamlines are generated by sampling from Gaussian noise conditioned on $\mathbf{z}$.}
\label{fig:fullpipeline}
\end{figure*}

We now introduce the architecture of GenTract, which comprises two main components: a global fODF encoder and a conditional Transformer trained using either a Diffusion or FM objective. These components, summarized in Figure~\ref{fig:fullpipeline}, jointly address the challenges outlined in the introduction. The encoder learns a global context embedding, which is used as a condition by the generator to sample all streamline coordinates simultaneously. Each streamline is thus produced as a whole rather than stepwise.

\subsection{Anatomical Conditioning Encoder}
The model's first task is to convert the high-dimensional fODF data into a compact and informative conditioning tensor $\mathbf{z}$. The encoder processes these volumes in a two-stage process:
\begin{enumerate}
    \item \textit{Per-Coefficient Representation Learning.} We train $m$ independent VAEs, each modeling a distinct SH coefficient volume (Figure~\ref{fig:fullpipeline}A). The $i$-th VAE includes an encoder $\mathcal{E}^{(i)}$ that maps the volume ($H, W, D$) to a compact latent space ($C_z, H_z, W_z, D_z$) and a decoder $\mathcal{D}^{(i)}$ that reconstructs it. Each VAE is trained using a composite loss function comprising reconstruction, perceptual, and adversarial terms, along with a Kullback-Leibler regularization term, following the formulation in~\cite{guo2025maisi}. This process yields a set of $m$ latent embeddings, $z^{(0)}, \dots, z^{(m-1)}$, that compactly represent the full fODF information.

    \item \textit{Fusion and Refinement.} Each $z^{(i)}$ is further refined by a 3D ResNet-like class-conditioned encoder $\mathcal{E}^c$, producing $\hat{z}^{(i)} = \mathcal{E}^c(z^{(i)}, i)$ of shape $(C_c, H_c, W_c,D_c)$. The weights of $\mathcal{E}^c$ are shared across all $m$ coefficients, while the index $i$ is provided as a conditioning signal to preserve coefficient-specific context. Additionally, $\mathcal{E}^c$ further reduces the spatial dimensionality of the latent representations to lower the overall computational cost. Unlike the VAEs, $\mathcal{E}^c$ is trained jointly with the downstream generative model, enabling it to adaptively extract the most relevant information for the tractography task. The spatial dimensions of each $\hat z^{(i)}$ are flattened, and all $m$ latents are concatenated along the channel dimension, resulting in the global conditioning tensor $\mathbf{z}$ of shape $(H_c \times W_c \times D_c, m \times C_c)$.
    
\end{enumerate}

\subsection{Conditional Transformer}
The core generative model is a Transformer~\cite{vaswani2017attention} designed to operate on batches of 3D coordinates representing streamlines. It is trained to learn the mapping from noise to a clean streamline, conditioned on the anatomical context tensor $\mathbf{z}$.

\paragraph{Architecture.} The model receives three inputs: (i) a noisy streamline $x_t$ of shape $(p, 3)$, where $p$ is the number of sampled points on the streamline; (ii) the timestep $t$; and (iii) the context tensor $\mathbf{z}$. To enable cross-attention between $\mathbf{z}$ and the streamline representation, these inputs are linearly projected into a shared embedding space of dimension $n$. Following~\cite{peebles2023scalable}, we add a sinusoidal positional encoding and a learned temporal embedding to the streamline representation. The main body of the model is a stack of $M$ Transformer layers containing two key attention mechanisms:
\begin{itemize}
    \item \textit{Self-Attention}: Models the geometric relationships and dependencies between points along the streamline itself, ensuring internal coherence.
    \item \textit{Cross-Attention}: Integrates guidance from the projected $n$-dimensional anatomical context $\mathbf{z}$, allowing the global white matter structure to inform the streamline's generation.
\end{itemize}
The final output from the Transformer stack is projected back to the original 3D coordinate space to compute the training loss.

\paragraph{Training Objectives.} The Transformer backbone is trained to learn the dynamics of the continuous time-process from Gaussian noise to target streamlines, conditioned on the learned $\mathbf{z}$, using either the Diffusion objective, $\mathcal{L}_{D}$, as described in \eqref{eq:dmeqn} (Figure \ref{fig:fullpipeline} B(1)), or the FM objective $\mathcal{L}_{FM}$, as in \eqref{eq:fm_loss} (Figure \ref{fig:fullpipeline} B(2)). In both cases, the loss is back-propagated through the Transformer network, as well as the class-conditioned encoder, $\mathcal{E}^c$. Given a global conditioning tensor $\mathbf{z}$, the training objective allows the model to learn the target distribution over streamlines. 

\subsection{Full Tractogram Inference}
At inference for a new subject (Figure \ref{fig:fullpipeline} (C)), we first compute the conditioning tensor $\mathbf{z}$ from the input SH coefficients. The model then generates streamlines by applying the learned reverse process, starting from random noise and conditioned on the subject-specific $\mathbf{z}$. Repeating this process in batches allows us to sample from the full conditional distribution of streamlines learned by the model for a specific subject. An example of a generated tractogram from GenTract can be seen in Figure \ref{fig:intro_figure}, along with further qualitative results in the Supplementary Material.

\subsection{Implementation Details}\label{sec:implementation}
We detail all the data dimensions and hyperparameters in the Supplementary Material. The VAEs were obtained by fine-tuning separate instances of the MAISI VAE~\cite{guo2025maisi}. The Transformer blocks, as well as training scripts and data processing, for both the Diffusion and FM models were implemented via MONAI~\cite{pinaya2023generative}. All experiments are performed on an NVIDIA H100 GPU~\cite{choquette2023nvidia}. All training and inference code can be found at \href{https://github.com/alecsargood/GenTract}{https://github.com/alecsargood/GenTract}.

%% file: sec/4_experiments.tex
\section{Experiments}
\label{sec:experiments}
This Section evaluates the GenTract framework. We first introduce the datasets and evaluation protocols. We then conduct an internal model comparison to determine GenTract's optimal configuration, examining the choice of generative framework, model size, and number of inference steps. Next, we provide a comparative benchmark of this optimal model against SOTA methods on high-quality data. Finally, we perform a robustness analysis, assessing all methods under noisy and low-resolution conditions using both internal and external test sets. We provide details on all statistical tests performed in the experiments in the Supplementary Material.

\subsection{Datasets and Preprocessing}\label{sec:data}

\paragraph{Training and Test Datasets.} We use dMRI data from 1,042 subjects (age 22-35, 45.6\% female) from the publicly available Human Connectome Project (HCP) Young Adult dataset~\cite{van2013wu}. For the supervised training target, we use the fODF and tractography data processed by~\cite{kruper2021evaluating}, which employs the PyAFQ tractography pipeline. PyAFQ first applies constrained spherical deconvolution to the raw dMRI data to estimate the fODF by computing its SH coefficients, and then generates a tractogram using probabilistic tractography~\cite{contributors2014dipy}. It then registers the subject's data to a template and uses its own filtering pipeline to filter streamlines based on geometric plausibility and alignment with a predefined white-matter atlas containing 24 known bundles (collection of streamlines that form neuroanatomical pathways connecting two regions of the brain)~\cite{kruper2021evaluating}. Each subject is thus represented by a pair: a 4D SH coefficient image and its corresponding filtered tractogram. We use a strict subject-level split for training (75\%), validation (10\%), and test (15\%) sets. The training set is augmented using deterministic rotations $(\pm15^{\circ}, \pm30^{\circ}, \pm45^{\circ})$ of the tractograms and their corresponding SH volumes. We use the HCP test set to establish baseline performance and then evaluate model robustness under several challenging conditions. First, we assess performance on noisy data by adding Rician noise ($\sigma=0.005$) to the input fODF signal of the HCP test set. Next, to simulate low-field data, we synthetically downsample the HCP test set to $3\text{ mm}^3$ isotropic resolution and similarly add Rician noise ($\sigma=0.005$), with resolution parameters aligned to previous studies of tractography on low-field MR scanners~\cite{gholam2025diffusion}. Finally, we assess GenTract’s out-of-domain performance in low-resolution (LR) and noisy conditions using the 28-subject test set from TractoInferno~\cite{poulin2022tractoinferno}, a highly heterogeneous dataset compiled from multiple studies (ages 18–62, with varying acquisition protocols) and extensively preprocessed to ensure research-grade quality. Specifically, we synthetically downsample this external dataset and introduce noise in the same manner as the HCP test set, enabling evaluation of robustness under LR, noisy, and cross-dataset conditions.

\paragraph{Data Preprocessing.}
To reduce computational load for training of the model, all SH coefficient volumes are downsampled to $1.875\text{ mm}^3$ isotropic resolution. Each of the SH coefficient volumes are \textit{z}-score normalized, and all streamline coordinates are min-max scaled to $[-1,1]$, using statistics computed from the training set.

\subsection{Evaluation}

\paragraph{Evaluation Metrics.} Evaluating tractograms is challenging as there is no absolute biological `ground truth', thus we must rely on proxies to assess model performance. As done in previous studies~\cite{legarreta2021filtering,theberge2021track}, we define a streamline as a True Positive (TP) if it is retained by a given filtering tool (the proxy for truth), and a False Positive (FP) if it is discarded. Given this, we evaluate all methods on three criteria:

\begin{enumerate}
    \item \textit{Precision (Percentage of Retained Streamlines)}: This is defined as $\frac{\text{TP}}{\text{TP} + \text{FP}}$. A high-precision score indicates that a large proportion of the generated streamlines are valid (TP), and that the model minimizes the creation of false positive streamlines (FP).
    
    \item \textit{Number of Discovered Bundles}: This metric assesses the model's ability to reconstruct a predefined set of bundles, using a white matter atlas as a reference. We count how many of these canonical bundles are successfully populated by plausible streamlines. This metric also serves as a proxy for streamline recall, where a missed bundle (and thus missed streamlines) represent false negatives (FNs).
    
    \item \textit{Inference Time}: We report the mean computation time required to generate a full tractogram for a test subject. 
\end{enumerate}

\paragraph{Internal Model Comparison Tools.} For the internal model comparisons, we evaluate tractograms using the PyAFQ filtering mechanism~\cite{kruper2021evaluating}. This is the same filtering mechanism used to filter the supervised training data, and we use this tool to calculate both precision (AFQ \% P) and number of recovered bundles (AFQ Bundles, up to 24). This provides a consistent metric to evaluate performance on the internal test set to determine the optimal configuration for GenTract.

\paragraph{SOTA Comparison Evaluation Tools.} To avoid any potential bias from using PyAFQ (which was used in training), the main SOTA comparison relies on two independent and established tools. The first, BundleSeg (BS)~\cite{st2023bundleseg}, provides a measure of precision (BS \% P) and number of recovered bundles (BS Bundles). BundleSeg operates by registering a subject's streamlines into a standard template space. There, streamlines are segmented against a high-resolution, population-based atlas (51 bundles) that serves as a comprehensive reference for `valid' pathways. This subject-specific registration penalizes any streamlines not well-aligned with the subject's unique anatomy. The second, TractOracle-Net (TO-Net)~\cite{theberge2024tractoracle}, serves as an independent measure of precision (TO-Net \% P). TO-Net is a pretrained transformer that classifies streamlines as plausible or implausible based on their geometric properties. Its diverse training provides an assessment of the model's output, independent of a single atlas.

\subsection{Internal Model Comparison Study}

\paragraph{Choice of Architecture and Generative Framework.}\label{sec:ablation1} We perform a comparison study by: (1) training GenTract using either the Diffusion or FM frameworks, (2) varying the number of Transformer layers ($M \in \{4, 6, 8\}$), and (3) evaluating different model embedding dimensions for the best-performing configuration. We do not consider transformer layers beyond $M=8$ because of computational expense. Inference for all architectures is conducted using 50 steps. The quantitative results are summarized in Table~\ref{tab:model_performance_unfiltered}. All model configurations successfully identify all 24 AFQ bundles. The primary differences appear in the AFQ \% P, where performance improves consistently with increasing numbers of layers for both architectures. Moreover, Diffusion models outperform FM models at each corresponding layer depth. The 8-layer Diffusion model ($M = 8$) achieves the highest precision at $85.0 \pm 1.5\%$, compared to $82.45 \pm 1.67\%$ for its FM counterpart. Finally, Table~\ref{tab:model_performance_unfiltered} identifies $n=256$ as the optimal embedding dimension for a fixed architecture, as smaller dimensions ($n=128$) underperform and larger dimensions ($n=512$) cause overfitting. We therefore select the Diffusion model framework, with $M=8$ and $n=256$, as the optimal GenTract architecture.

\begin{table}[h!]
\centering
\caption{Performance summary of tractography generation models. The table shows the number of bundles found and the percentage of streamlines retained. Values are mean $\pm$ std. Best performing results in each column in \textbf{bold} and second best \underline{underlined}.}
\label{tab:model_performance_unfiltered}
\resizebox{\linewidth}{!}{
\begin{tabular}{@{}lccc@{}}
\toprule
\textbf{Model Type} & $\textbf{M}$ & \textbf{AFQ Bundles ( / 24)} & \textbf{AFQ \% P} \\ \midrule
Diffusion & 4 & $24.0 \pm 0.0$ & $81.8 \pm 1.5$\\
& 6 & $24.0 \pm 0.0$ & $\underline{83.4 \pm 1.7}$\\
& 8 & $24.0 \pm 0.0$ & $\mathbf{85.0 \pm 1.5}$ \\ \midrule
Flow Matching & 4 & $24.0 \pm 0.0$ & $75.50 \pm 2.14$\\
& 6 & $24.0 \pm 0.0$ & $80.56 \pm 1.78$\\
& 8 & $24.0 \pm 0.0$ & $82.45 \pm 1.67$\\ \toprule
Diffusion& & &\\
$n = 128$ & 8 & $24.0 \pm 0.0$ & $80.6 \pm 1.1$ \\
$n = 256$ & 8 & $24.0 \pm 0.0$ & $\mathbf{85.0 \pm 1.5}$ \\
$n = 512$ & 8 & $24.0 \pm 0.0$ & $79.3 \pm 1.5$ \\
\bottomrule
\end{tabular}
}
\end{table}

\begin{figure}[h!]
    \centering
    % Use 0.6\linewidth to make it smaller than the full text width
    \includegraphics[width=0.85\linewidth]{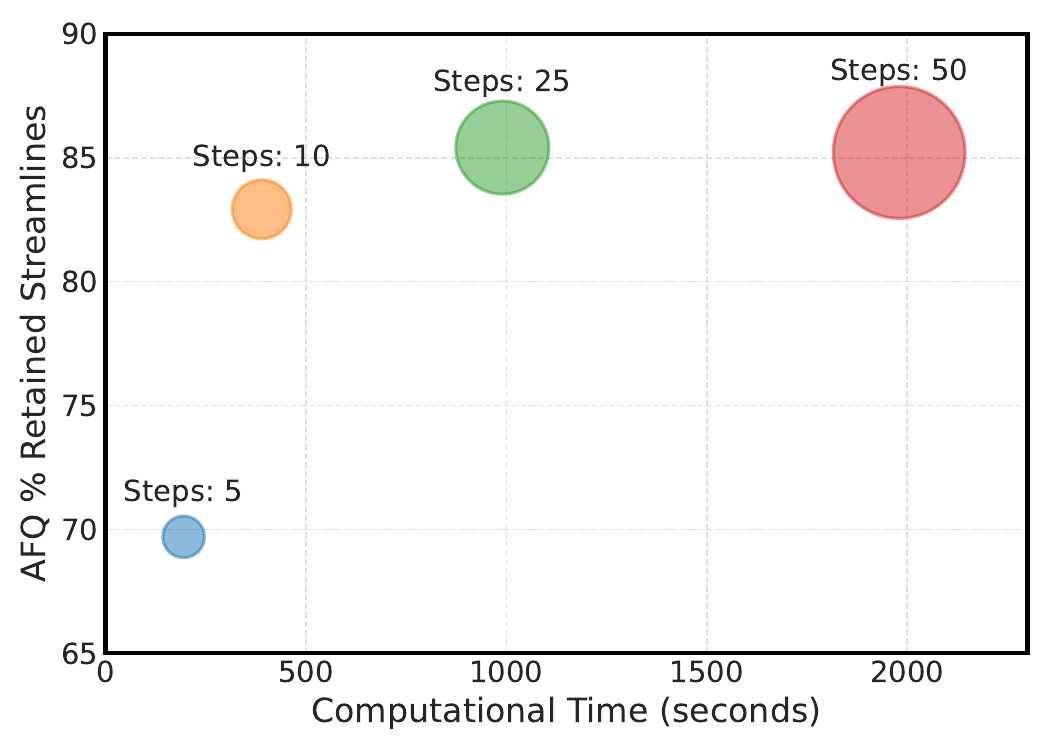}
    \caption{AFQ streamlines retained (AFQ \% P) vs. Computational Time by Inference Step for the Diffusion ($M=8$, $n=256$) model. The size of each bubble is proportional to the number of DDIM inference steps used.}
    \label{fig:inference_tradeoff}
\end{figure}

\paragraph{Number of Inference Steps.}\label{sec:ablation3} Figure~\ref{fig:inference_tradeoff} illustrates the trade-off between precision (AFQ \% P) and computational time across different numbers of Denoising Diffusion Implicit Models (DDIM) inference steps. We observe a substantial performance jump from 5 steps (69.7\% AFQ P) to 10 steps (82.9\% AFQ P), but we note that gains diminish thereafter. The 25-step (85.4\%) and 50-step (85.2\%) configurations offered only small precision improvements while more than doubling the inference time of the 10-step model (from 391s to 1976s). All variations recovered 24 bundles. We therefore select 10 DDIM inference steps for all subsequent experiments.

\subsection{Comparison with State-of-the-Art}

\paragraph{Baselines.}\label{sec:baselines}
We benchmark the optimal GenTract configuration against classical local tractography methods (iFOD2~\cite{tournier2010improved} and SD Stream~\cite{descoteaux2008deterministic}), deep learning–based local methods (TractOracle~\cite{theberge2024tractoracle} and DDTracking~\cite{li2025ddtracking}), and a standard global method (tckglobal~\cite{christiaens2015global}). The iFOD2, SD Stream, and tckglobal baselines are implemented using MRtrix3~\cite{tournier2019mrtrix3} and tractograms generated by iFOD2 and SD Stream are further processed with SIFT filtering~\cite{smith2013sift}, following standard practice. For TractOracle and DDTracking, we used the pretrained models provided in their open-source implementations, as this empirically yielded better performance than retraining. Further details of all inference pipelines settings are in the Supplementary Material.

\paragraph{Quantitative Comparison: Base Performance.} Table~\ref{tab:baseline_performance} presents the quantitative comparison on the HCP test set, which highlights a clear trade-off between streamline precision (minimizing FPs) and bundle count (minimizing FNs). GenTract achieves a state-of-the-art precision: we observe that on the BS \% P metric, GenTract (61.95\%) outperforms the next-best deep learning methods, DDTracking (35.20\%) and TractOracle (28.93\%), by a factor of 1.8 and 2.1, respectively. This advantage is consistent on the TO-Net \% P metric, where GenTract (56.35\%) again surpasses DDTracking (30.70\%) and TractOracle (39.55\%). This gain in precision comes at the cost of a higher false negative count, as the other methods identify more bundles than GenTract (36.62). While GenTract has a higher FN rate than the baselines, we hypothesize that this is a consequence of training on a constrained 24-bundle (PyAFQ) distribution. 

\begin{table}[h!]
\centering
\caption{Comparison of tractography models on HCP test set. Best performance is \textbf{bolded} and second-best is \underline{underlined} (mean $\pm$ std).}
\label{tab:baseline_performance}
\resizebox{\linewidth}{!}{
\begin{tabular}{@{}lccc@{}}
\toprule
\textbf{Model} & \textbf{BS \% P} & \textbf{BS Bundles (/51)} & \textbf{TO-Net \% P} \\
\midrule
tckglobal & $0.19 \pm 0.05$ & $42.91 \pm 2.75$ & $17.83 \pm 1.03$\\
iFOD2 & $1.96 \pm 0.31$ & \underline{48.88 $\pm$ 1.49} & $6.30 \pm 0.60$\\
SD Stream & $4.71 \pm 0.67$ & $47.85 \pm 1.63$ & $11.10 \pm 0.80$\\
DDTracking & \underline{35.20 $\pm$ 3.49} & \textbf{49.69 $\pm$ 0.64} & $30.70 \pm  6.00$\\
TractOracle & $28.93 \pm 3.28$ & $48.20 \pm 0.72$ & \underline{39.55 $\pm$ 1.99}\\
GenTract & \textbf{61.95 $\pm$ 4.50} & $36.62 \pm 1.21$ & \textbf{56.35 $\pm$ 4.25}\\
\bottomrule
\end{tabular}
}
\end{table}

\paragraph{Quantitative Comparison: Robustness to Rician Noise.} 
Table~\ref{tab:noise_robustness} summarizes the quantitative comparison under Rician noise, as described in Section~\ref{sec:data}. We observe that iFOD2, SD Stream, and DDTracking are severely impacted by noise, with precision rates falling substantially and the number of bundles recovered dropping. TractOracle exhibits moderate robustness, with its BS \% P decreasing by 23.8\%, though it retains the highest bundle count. Tckglobal remains relatively stable under noise, still reconstructing many bundles, but remains uncompetitive in terms of precision. In contrast, GenTract shows strong stability across all metrics, with only a 2.6\% decrease in BS \% P, 3.8\% in BS Bundles, and 3.3\% in TO-Net \% P, and although it has a higher FN rate than tckglobal and TractOracle, it maintains the highest precision among all baseline methods. These findings indicate that GenTract preserves high streamline precision under noisy conditions, showing its potential as a promising choice for challenging diffusion MRI data present in clinical scenarios. 

\begin{table}[h!]
\centering
\caption{Comparison of tractography models under Rician noise. Best performance is \textbf{bolded} and second-best is \underline{underlined} (mean $\pm$ std).}
\label{tab:noise_robustness}
\resizebox{\linewidth}{!}{
\begin{tabular}{@{}lccc@{}}
\toprule
\textbf{Model} & \textbf{BS \% P} & \textbf{BS Bundles (/51)} & \textbf{TO-Net \% P} \\
\midrule
tckglobal & $0.20 \pm 0.05$ & \underline{43.11 $\pm$ 2.84} & $17.85 \pm 1.05$ \\
iFOD2 & $0.08 \pm 0.04$ & $13.17 \pm 2.22$ & $5.03 \pm 0.73$ \\
SD Stream & $0.11 \pm 0.04$ & $11.40 \pm 2.60$ & $9.53 \pm 0.92$ \\
DDTracking & $5.41 \pm 1.84$ & $39.43 \pm 2.60$ & $12.24 \pm 1.95$ \\
TractOracle & \underline{22.06 $\pm$ 3.08} & \textbf{48.29 $\pm$ 0.59} & \underline{34.90 $\pm$ 2.35} \\
GenTract & \textbf{60.32 $\pm$ 4.12} & $35.20 \pm 1.43$ & \textbf{54.52 $\pm$ 4.02} \\
\bottomrule
\end{tabular}
}
\end{table}

\paragraph{Quantitative Comparison: Robustness to Low-Resolution and Noise.}\label{sec:lowrescomp} 
Table \ref{tab:low_res_performance} summarizes the results on the low-resolution and noisy HCP dataset, as described in Section~\ref{sec:data}. While an absolute drop is expected for all methods on degraded data, we observe that GenTract retains a higher level of relative performance, while baseline methods degrade more severely. Local methods (iFOD2, SD Stream) struggle in this setting, with BS \% P dropping to 0\%. This behavior is consistent with their autoregressive design, which is highly sensitive to ambiguous local information and prone to error propagation. The global method (tckglobal) also fails completely. This is likely due to its classical optimization which forms incorrect streamline connections under low-resolution ambiguity, leading to a collapse in precision. Although TractOracle continues to produce the highest number of bundles, its BS \% P rate decreases to nearly 1\%, and TO-Net \% P to less than 6\%. DDTracking similarly has low-precision scores. In contrast, GenTract still achieves a BS \% P of 15.73\%, an order of magnitude higher than the next-best result, and a TO-Net \%-P of 43.43\%, while also recovering the second-highest bundle count. These results suggest that GenTract's performance advantage is maintained on this low-resolution and noisy data.

\begin{table}[h!]
\centering
\caption{Comparison of tractography models under low-resolution and noise. Best performance is \textbf{bolded} and second-best is \underline{underlined} (mean $\pm$ std).}
\label{tab:low_res_performance}
\resizebox{\linewidth}{!}{
\begin{tabular}{@{}lccc@{}}
\toprule
\textbf{Model} & \textbf{BS \% P} & \textbf{BS Bundles (/51)} & \textbf{TO-Net \% P} \\
\midrule
tckglobal & $0.00 \pm 0.00$ & $0.00 \pm 0.00$ & $3.62 \pm 1.30$ \\
iFOD2 & $0.00 \pm 0.00$ & $0.18 \pm 0.39$ & $0.55 \pm 0.34$ \\
SD Stream & $0.00 \pm 0.00$ & $0.00 \pm 0.00$ & $3.40 \pm 2.51$ \\
DDTracking & \underline{4.44 $\pm$ 1.45} & $31.71 \pm 3.03$ & \underline{8.42 $\pm$ 0.60} \\
TractOracle & $1.12 \pm 0.30$ & \textbf{37.79 $\pm$ 2.45} & $5.41 \pm 0.73$ \\
GenTract & \textbf{15.73 $\pm$ 2.82} & \underline{31.74 $\pm$ 1.26} & \textbf{43.43 $\pm$ 3.68}\\
\bottomrule
\end{tabular}
}
\end{table}

\paragraph{Quantitative Comparison: External Robustness to Low-Resolution and Noise.}\label{sec:extlowrescomp} Table~\ref{tab:external_low_res} presents the external robustness evaluation, assessing GenTract's generalization on the synthetically-degraded TractoInferno dataset. We observe that most models perform better on this dataset than on the degraded HCP data, which we hypothesize is due to the better underlying image quality of the TractoInferno dataset as a result of its extensive preprocessing. Despite this general uplift, the performance hierarchy remains clear. Classical and global methods (iFOD2, SD Stream, tckglobal) still fail completely. DDTracking shows only modest TO-Net \% P (9.79\%). TractOracle, while recovering the most bundles (42.18), achieves a limited precision of 9.74\% BS \% P. In contrast, GenTract achieves the highest precision by a wide margin (24.94\% BS \% P and 40.88\% TO-Net \% P) and the second-highest bundle count (35.05). These results suggest that GenTract retains its performance advantage on this low-quality, out-of-domain data, indicating its design generalizes beyond the training domain.

\begin{table}[h!]
\centering
\caption{Performance on low-resolution and noisy TractoInferno dataset. Best performance is \textbf{bolded} and second-best is \underline{underlined} (mean $\pm$ std).}
\label{tab:external_low_res}
\resizebox{\linewidth}{!}{
\begin{tabular}{@{}lccc@{}}
\toprule
\textbf{Model} & \textbf{BS \% P} & \textbf{BS Bundles (/51)} & \textbf{TO-Net \% P} \\
\midrule
tckglobal & $0.00 \pm 0.00$ & $0.00 \pm 0.00$ & $2.24 \pm 0.68$ \\
iFOD2 & $0.01 \pm 0.02$ & $0.36 \pm 0.49$ & $1.34 \pm 0.89$ \\
SD Stream & $0.01 \pm 0.04$ & $0.18 \pm 0.39$ & $4.24 \pm 2.75$ \\
DDTracking & $6.35 \pm 7.73$ & $17.96 \pm 9.89$ & $9.79 \pm 1.44$\\
TractOracle & \underline{9.74 $\pm$ 3.53} & \textbf{42.18 $\pm$ 3.38} & \underline{10.34 $\pm$ 1.34} \\
GenTract & \textbf{24.94 $\pm$ 8.09} & \underline{35.04 $\pm$ 1.54} & \textbf{40.88 $\pm$ 5.01} \\
\bottomrule
\end{tabular}
}
\end{table}

\begin{figure}[h!]
  \centering
  \includegraphics[width=0.9\linewidth]{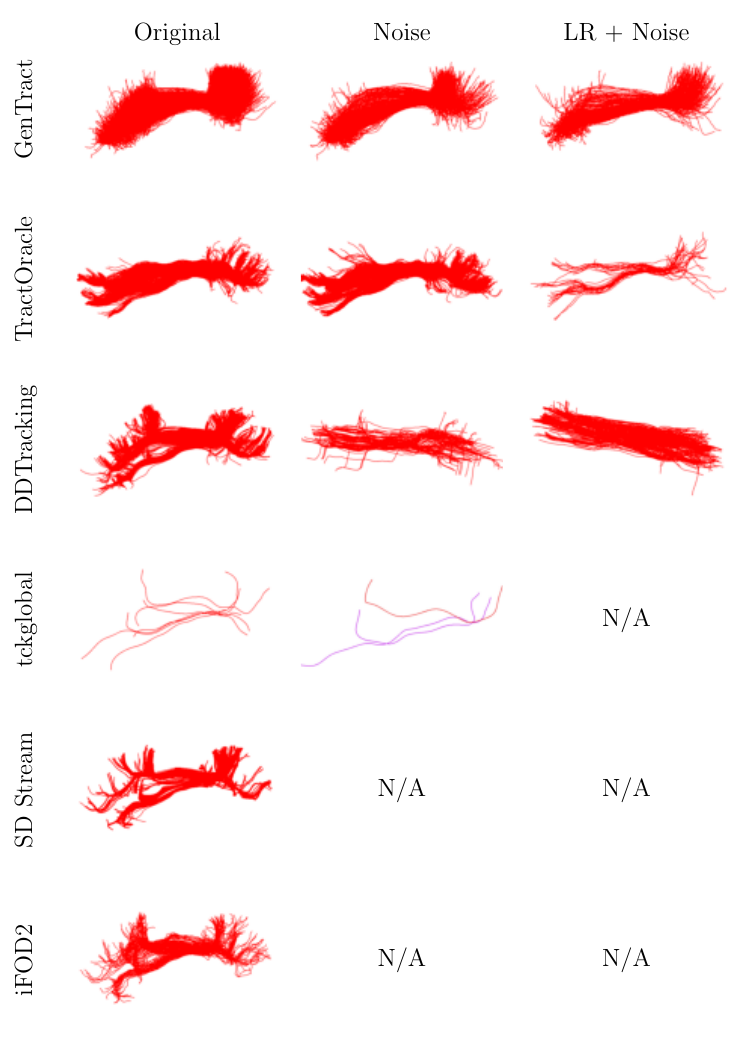}
  \caption{Qualitative result showing SLFR segmented by BundleSeg for each method, on original, noised, and LR+Noise for a HCP test subject.}
  \label{fig:qualitative}
\end{figure}

\paragraph{Qualitative Comparison.} Figure~\ref{fig:qualitative} illustrates the Right Superior Longitudinal Fasciculus (SLFR) bundle for a single HCP test subject (segmented by BundleSeg), for all methods under different data conditions. GenTract has a higher streamline retention compared with baseline methods under the Noise and LR + Noise settings. This supports the quantitative performance differences we report in the previous tables. SD Stream, iFOD2, and tckglobal fail to produce any valid streamlines for this bundle in LR + Noise setting. This aligns with their low bundle counts and precision scores in the quantitative analysis. Further qualitative results can be found in the Supplementary Material.

\paragraph{Inference Times.} Figure \ref{fig:comp} shows that GenTract is 2-3$\times$ faster than classical methods and achieves inference times comparable to other deep learning methods. 

%% file: sec/5_discussion.tex
\section{Discussion}
\label{sec:discussion} In this study, we present GenTract, a new paradigm that frames global tractography as a conditional generative task. GenTract addresses limitations of prior methods: it avoids step-wise error propagation of local approaches by generating all streamline coordinates simultaneously. Moreover, this generative formulation removes the need for initial seeding points, a source of operator-dependent variability in local tracking. We hypothesize that leveraging global conditioning from the entire dMRI contributes to its robustness against local noise and data perturbations. These properties show GenTract's potential for clinical applications, where scans are often time-constrained and of lower quality, supporting the practical adoption of global tractography.

\paragraph{Future Work.} Our study highlights important directions for future research. A primary challenge is the absence of ground truth in tractography, which gives rise to two related issues. First, there is potentially suboptimal supervision for GenTract. The results suggest that the specific choice of the PyAFQ filtered dataset leads to learning a constrained bundle distribution. While we train a high-precision model, it comes at a cost of a higher false negative rate. Future work should explore strategies to mitigate this trade-off, perhaps by expanding the distribution of streamlines present in the training set, or by developing global training approaches that do not require supervised data. Second, since no biological ground truth exists for tractography, we must rely on proxy evaluation tools that may have inaccuracies or introduce their own biases. 
A further limitation is the reliance on synthetic degradation for the robustness analysis. While the results on this data suggests that GenTract is more robust to noisy and low-resolution settings, a next step is to validate this performance on real-world clinical scans. Assessing its practical utility will require evaluating performance on non-synthetically degraded clinical and low-resolution data. Additionally, as we transition toward clinical viability, future work can exploit the probabilistic sampling nature of GenTract to quantify uncertainty in the generated tractograms.

\section*{Acknowledgements}
AS is supported by the EPSRC-funded UCL Centre for Doctoral Training i4health (EP/S021930/1) and by the Wellcome Trust CU-MONDAI project (WT 221915). LP is supported by the PNRR initiative (DM 118/2023). ET is supported by the Wellcome Trust CU-MONDAI project (WT 221915). DCA acknowledges funding from the E-DADS project (EU-JPND 2019; UKRI MR/T046422/1), the Wellcome Trust CU-MONDAI project (WT 221915), the Wellcome Trust Democratizing MRI project (WT 317797), and the NIHR UCLH Biomedical Research Centre.

%% file: sec/X_suppl.tex
\clearpage
\setcounter{page}{1}
\maketitlesupplementary

\section{Statistical Analysis}
\label{sec:stat}
Significance ($p<0.01$) was determined using paired t-tests with one-sided bootstrapping on the test set. 

\paragraph{Architecture (Table 1)} Diffusion models significantly outperformed FM models at each architectural size ($M=4, 6, 8$). Furthermore, the Diffusion model with $M=8$ and dimension $n=256$ yielded statistically superior results compared to all other tested architecture depths ($M=4, 6$) and widths ($n=128, 512$).

\paragraph{Benchmarking (Tables 2--5)} GenTract demonstrated statistically significant improvements over the second-best performing methods (TractOracle or DDTracking) in both BS \% P and TO-Net \% P across all experiments.

\section{Further GenTract Implementation Details}

\paragraph{Input representation and preprocessing.}
Following standard practice~\cite{theberge2024tractoracle,li2025ddtracking}, we use an SH order of $L_{\text{max}}=6$, resulting in $m=28$ SH coefficients. All streamlines are resampled to 128 points as in previous work~\cite{theberge2024tractoracle}. Each of the 28 SH 3D coefficient volumes is padded to shape $(H,W,D)=(88,112,88)$.

\paragraph{Latent spaces and architecture.}
We use latent dimension values of $(C_z,H_z,W_z,D_z)=(4,22,28,22)$ for the initial VAEs, and $(C_c,H_c,W_c,D_c)=(32,11,14,11)$ for the class-conditioned encoder $\mathcal{E}^c$. Within $\mathcal{E}^c$, coefficient-index conditioning ($i \in [0,m-1]$) is implemented via a learned embedding injected into the residual down-sampling blocks using the same additive mechanism commonly used for diffusion timestep embeddings. In the Transformer backbone, 8 attention heads are used for both self- and cross-attention.

\paragraph{Training setup and computational cost.}
All GenTract models are trained with a batch size of 1024 streamlines for 100 epochs. We use an initial learning rate of $5\times10^{-5}$ with cosine annealing over the full training cycle to a final learning rate of $1.25\times10^{-5}$. Training proceeds in two stages: representation learning, where each of the $m$ independent VAEs (one per SH coefficient) requires approximately 48 GPU-hours, and generative modeling, where the conditional Transformer requires 168--336 GPU-hours depending on the architecture.

\begin{figure}[h!]
    \centering
    \includegraphics[width=0.9\linewidth]{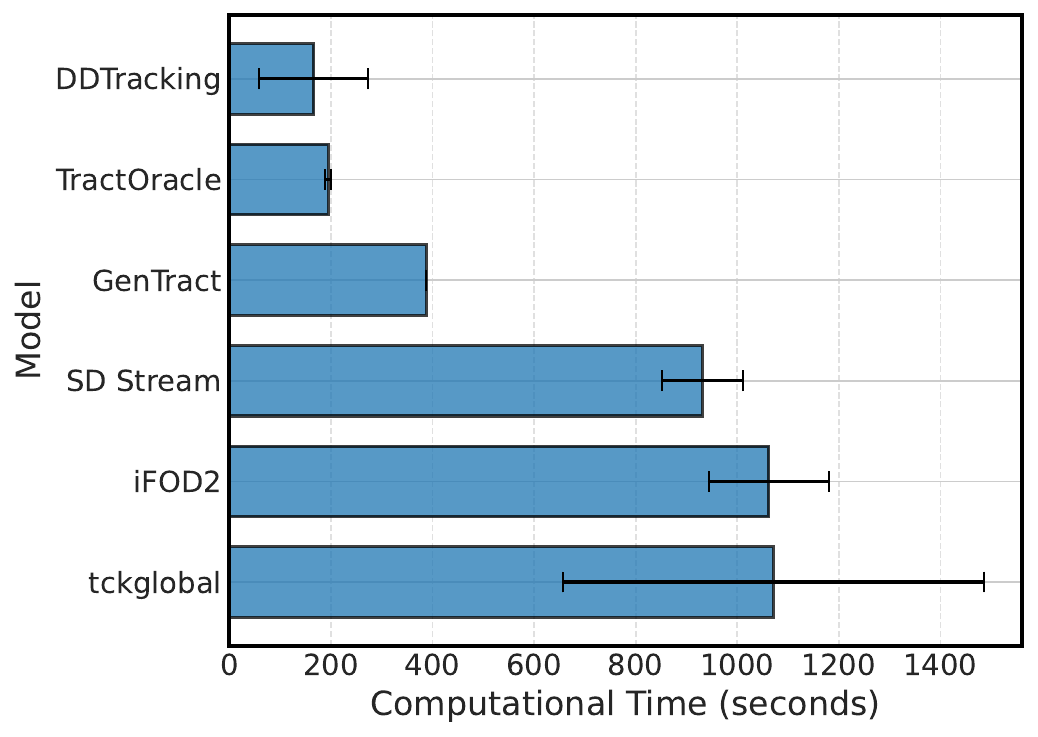}
    \caption{Computational time comparison for all methods. Bars show inference time (seconds) (mean $\pm$ std). The std for GenTract is very small.}
    \label{fig:comp}
\end{figure}

\section{Baseline Implementation Details}

For each of the DL-based methods (TractOracle, DDTracking), we use the standard inference pipelines from their open source implementations. Specifically, we run TractOracle inference with parameters: noise level: 0.1, seeds per voxel: 10, number of actors: 2500, and DDTracking with parameters: step size: 1, max angle: 45, max streamline length: 200, min streamline length: 40. Seeding masks are generated by dilating Fractional Anisotropy (FA) masks at FA $>$ 0.2 to approximate a white matter mask, which is then dilated (and subtracted) to form a White Matter / Gray Matter interface mask, as done in~\cite{theberge2024tractoracle}. For the classic non-DL methods (iFOD2, SD Stream, tckglobal), we use recommended practices for their inference~\cite{tournier2019mrtrix3}. Specifically, iFOD2 and SD Stream are used to generate $5\times10^6$ streamlines before undergoing SIFT filtering to $500\times 10^3$ streamlines, with seeding masks generated from 5ttgen~\cite{smith2012anatomically}. tckgen is used with number of iterations: $1\times 10^9$.

\section{Qualitative Results}

\begin{figure*}[h!]
  \centering
  \includegraphics[width=\linewidth]{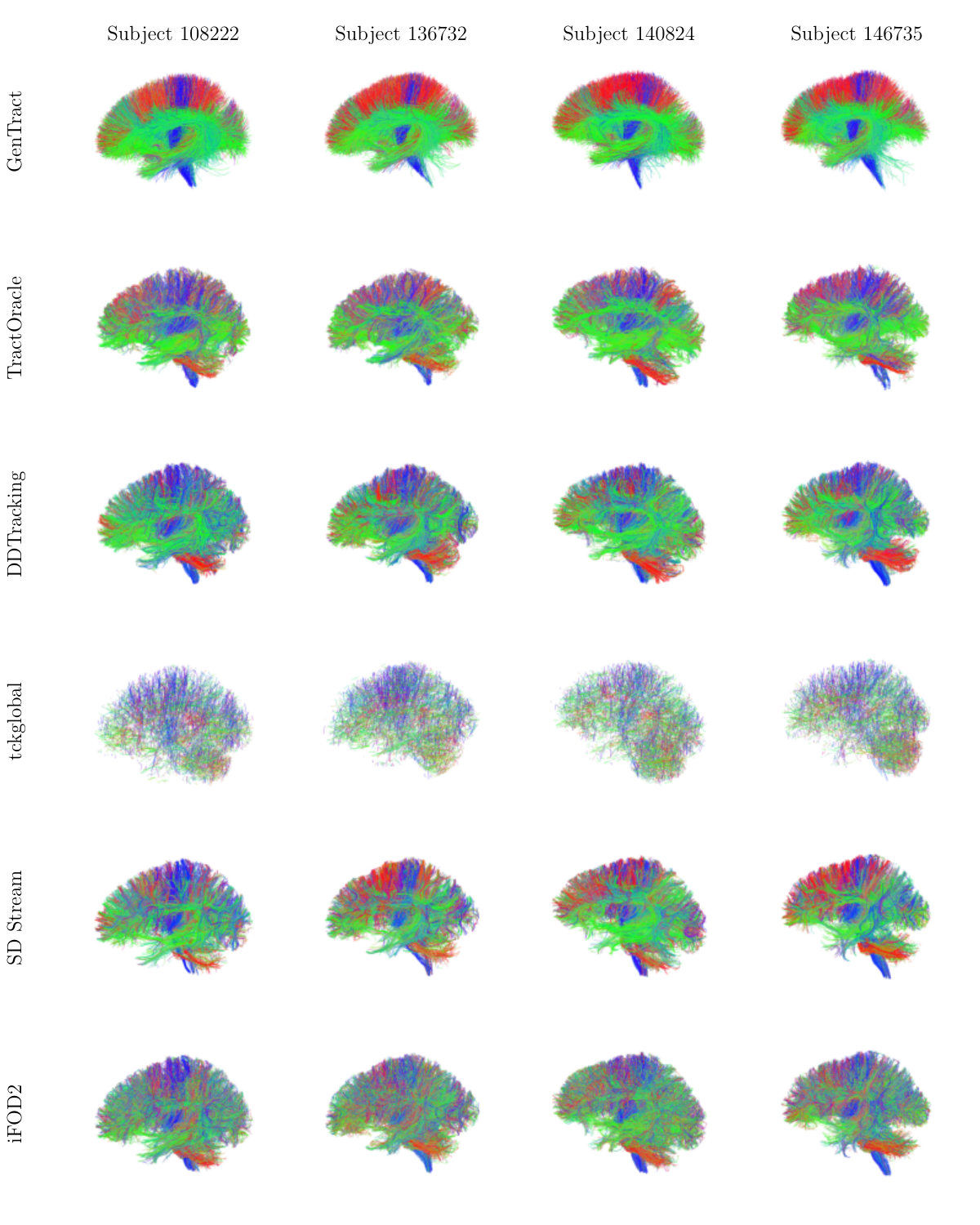}
  \caption{Qualitative result showing generated tractograms for 4 HCP test subjects for all methods on original (uncorrupted) data.}
  \label{fig:full_tractograms}
\end{figure*}

\begin{figure*}[h!]
  \centering
  \includegraphics[width=\linewidth]{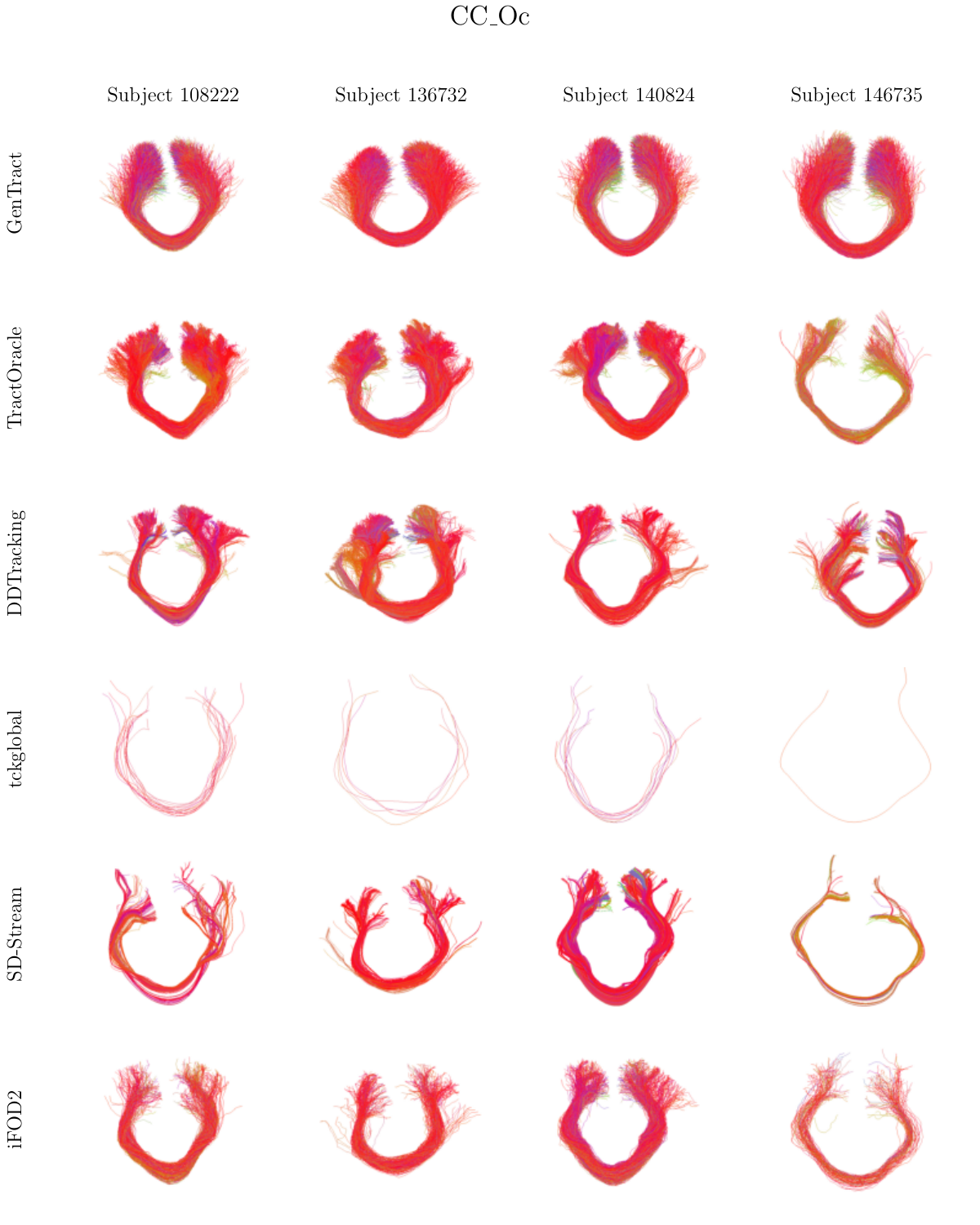}
  \caption{Qualitative result showing generated and segmented CC$\_$Oc bundles for 4 HCP test subjects for all methods on original (uncorrupted) data.}
  \label{fig:inter_CC}
\end{figure*}

\begin{figure*}[h!]
  \centering
  \includegraphics[width=\linewidth]{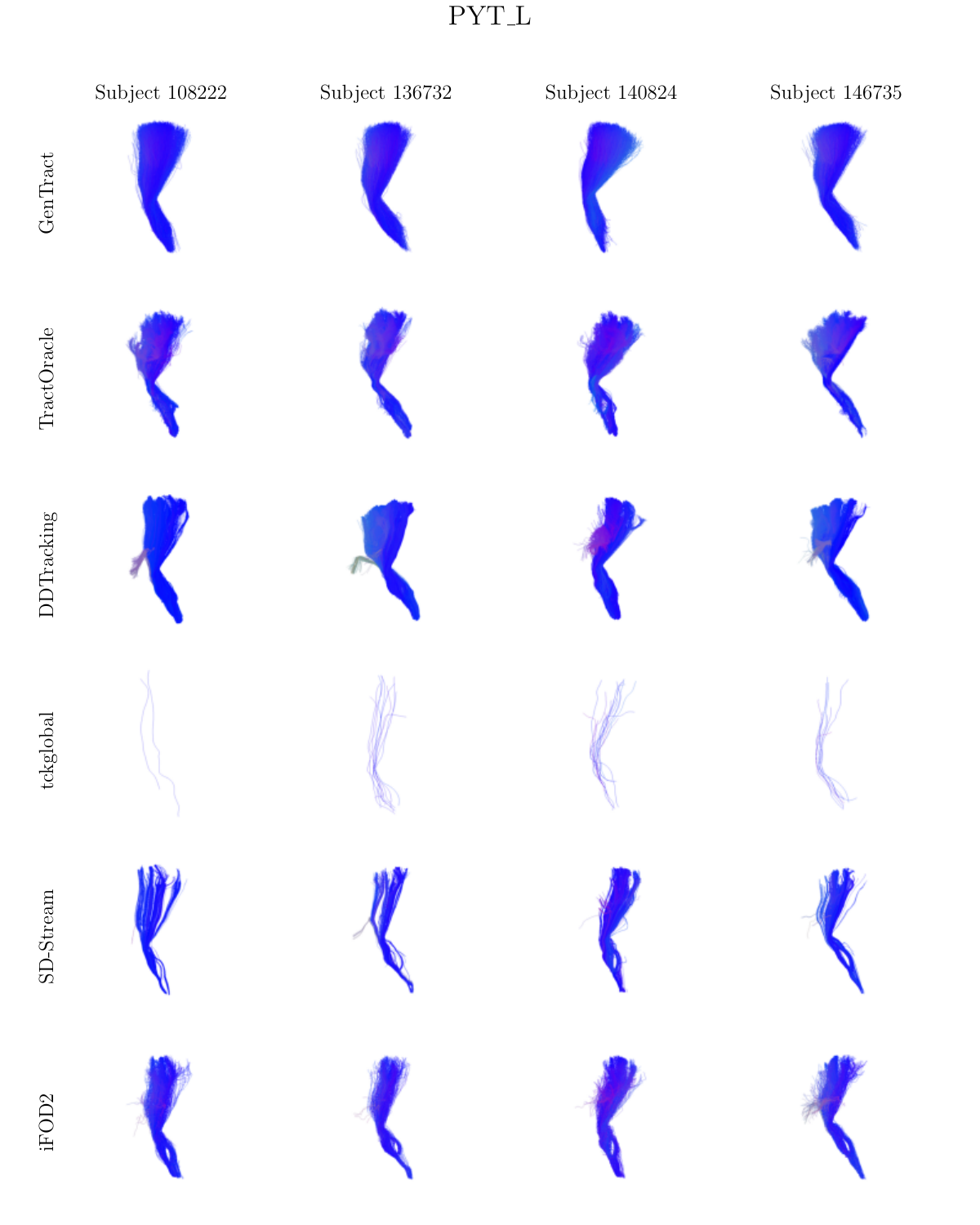}
  \caption{Qualitative result showing generated and segmented PYT$\_$L bundles for 4 HCP test subjects for all methods on original (uncorrupted) data.}
  \label{fig:inter_PYT}
\end{figure*}

\begin{figure*}[h!]
  \centering
  \includegraphics[width=\linewidth]{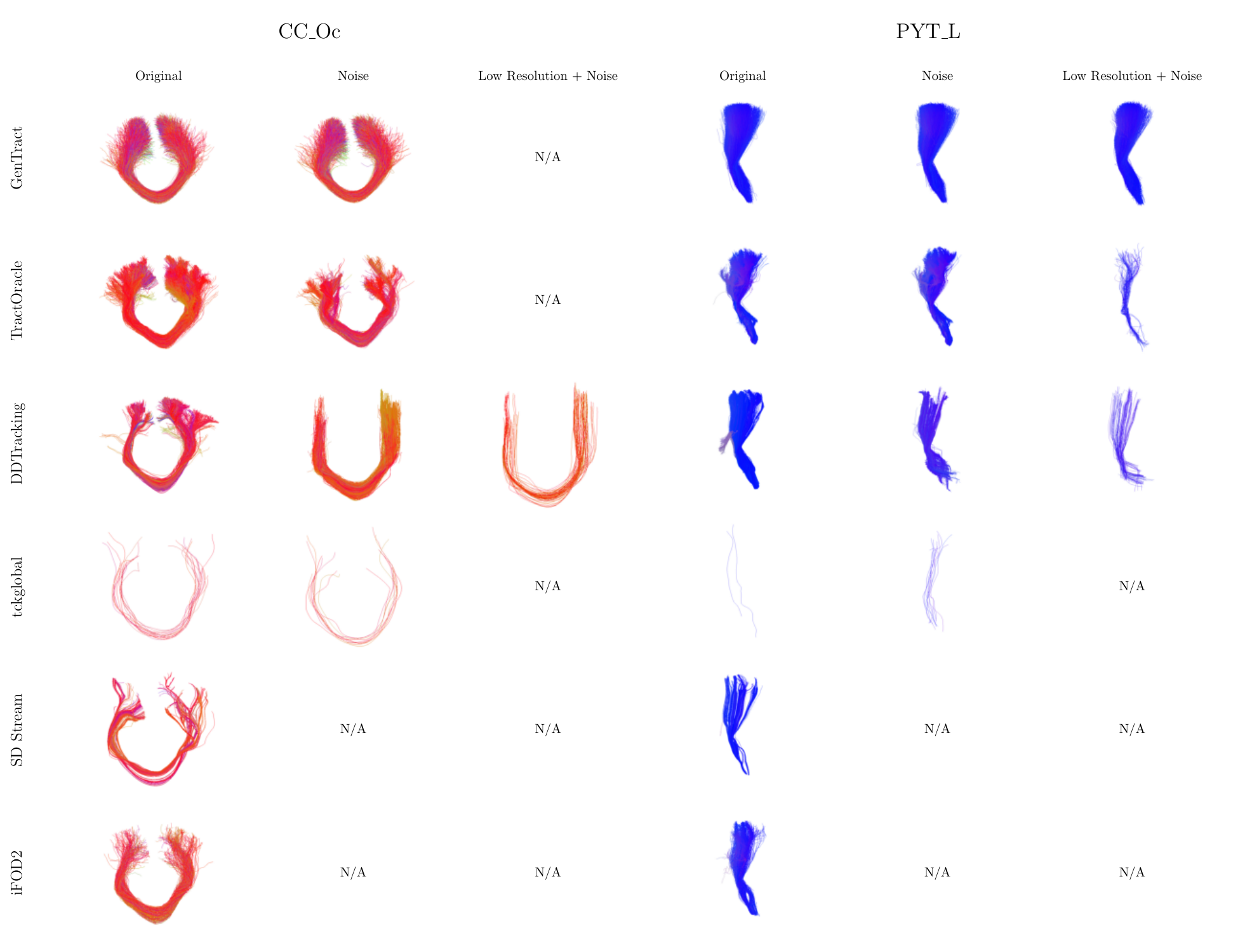}
  \caption{Qualitative result showing generated and segmented CC$\_$Oc and PYT$\_$L bundles for a single HCP test subject (108222) for all methods on original, noisy, and low-resolution and noise settings.}
  \label{fig:noisyfig}
\end{figure*}

In Figure \ref{fig:full_tractograms} we provide additional qualitative results to show generated tractograms (with 5,000 sampled streamlines) from GenTract, along with other baseline methods. We also show two different segmented bundles: the Corpus Callosum Occipital (CC$\_$Oc) in Figure \ref{fig:inter_CC}, and the Left Pyramid Tract (PYT$\_$L) in Figure \ref{fig:inter_PYT}, generated by GenTract across different subjects, showing subject-specific geometries. Additionally, we compare bundle-specific results across baseline methods to further assess sensitivity of these methods to noisy and low-resolution conditions (Figure \ref{fig:noisyfig}). In this figure, we present a failure mode of GenTract. We see that GenTract shows enhanced bundle attenuation under low-resolution and noisy conditions compared with baseline methods for the PYT$\_$L bundle, but similar to the other baseline methods, it is unable to resolve the CC$\_$Oc bundle.

